\documentclass[letterpaper, 10 pt, conference]{ieeeconf}  

\IEEEoverridecommandlockouts                              

\usepackage[caption=false,font=footnotesize]{subfig}
\usepackage{graphics} 
\usepackage{epsfig} 
\usepackage{mathptmx} 
\usepackage{times} 
\usepackage{amsmath} 
\usepackage{mathtools}
\usepackage{amssymb}  
\usepackage{xcolor}
\usepackage{graphicx}
\usepackage{subcaption} 
\usepackage{booktabs}
\usepackage{subcaption} 
\usepackage{siunitx}
\usepackage{xcolor}
\title{\LARGE \bf
Deep Reinforcement Learning for Autonomous Control of Time-Varying Accelerator Beamline Dynamics
}
\title{\LARGE \bf
Improved Robustness of Deep Reinforcement Learning for Control of Time-Varying Systems by Bounded Extremum Seeking
}

\author{Shaifalee Saxena$^{1, 2}$, Alan Williams$^{2}$, Rafael Fierro$^{1}$, Alexander Scheinker$^{2}$
\thanks{This work was supported by the U.S. Department of Energy (DOE),
Office of Science, Office of High Energy Physics contract number
9233218CNA000001 and the Los Alamos National Laboratory LDRD
Program Directed Research (DR) project 20220074DR. 
R. Fierro’s work is sponsored by Air Force Research Laboratory (AFRL) under agreements FA9453-18-2-0022 and FA9550-22-1-0093. Any opinions, findings, and
conclusions or recommendations expressed in this material are those of the authors and do not necessarily reflect the views of the United States Air Force.}
\thanks{$^{1}$Shaifalee Saxena and Rafael Fierro are with Department of Electrical and Computer Engineering, University of New Mexico, Albuquerque, NM 87106, USA,
        {\tt\small \{shaifalisaxena, rfierro\}@unm.edu}}%
\thanks{$^{2}$Shaifalee Saxena, Alan Williams and Alexander Scheinker are with Los Alamos National Lab, Los Alamos, NM 87547, USA
        {\tt\small \{shaifalees, awilliams, ascheink\}@lanl.gov}}%
}

\begin{document}

\maketitle
\thispagestyle{empty}
\pagestyle{empty}

\begin{abstract}
In this paper, we study the use of robust model-independent bounded extremum seeking (ES) feedback control to improve the robustness of deep reinforcement learning (DRL) controllers for a class of nonlinear time-varying systems. DRL has the potential to learn from large datasets to quickly control or optimize the outputs of many-parameter systems, but its performance degrades catastrophically when the system model changes rapidly over time. Bounded ES can handle time-varying systems with unknown control directions, but its convergence speed slows down as the number of tuned parameters increases and, like all local adaptive methods, it can get stuck in local minima. We demonstrate that together, DRL and bounded ES result in a hybrid controller whose performance exceeds the sum of its parts with DRL taking advantage of historical data to learn how to quickly control a many-parameter system to a desired setpoint while bounded ES ensures its robustness to time variations. We demonstrate the generality of our combined ES-DRL controller approach with numerical studies of three very different dynamic systems: 1) a general time-varying system, 2) automatic tuning of the Low Energy Beam Transport section at the Los Alamos Neutron Science Center linear particle accelerator, and 3) an intermittent-contact robotic block pushing task with a time-varying goal.
\end{abstract}

\section{Introduction}

Deep reinforcement learning (DRL) combines aspects of optimal control theory with deep learning techniques. In DRL, deep neural networks parameterize the policy (controller), the value function, or both. DRL is based on dynamic programming, introduced by Bellman to solve sequential decision problems through the Bellman optimality principle \cite{Bellman_1957}. Classical DP assumes known analytic models of the system dynamics and the reward function, which enables the synthesis of controllers $u$ that maximize cumulative return. Reinforcement learning (RL) relaxes this modeling requirement by learning from data collected as state, action, and reward transitions when the dynamics are unknown \cite{sutton1998reinforcement}. DRL extends RL to high-dimensional state and action spaces by using deep neural networks to approximate value functions and/or policies. This enables the application of RL to complex systems such as image-based perception and continuous control, where tabular methods become intractable \cite{arulkumaran2017deep}. A landmark example is deep Q-learning, which approximates the optimal action value function that maximizes cumulative reward \cite{mnih2015human}. DRL is being actively investigated for a broad set of applications, including robotic control \cite{tang8deep}, particle accelerator tuning \cite{hirlaender2020model}, and the fine-tuning of deep neural networks \cite{ouyang2022training}.

Handling time-varying systems remains a central challenge in DRL and is an active area of research \cite{ref:ContextRL}. When the system dynamics or reward function change substantially, the learned neural networks require retraining. Ongoing efforts aim to speed up learning from a few observations \cite{ref:FasterRL1,ref:FasterRL3} and to model partially observable RL problems for increased robustness \cite{ref:RLPartObs}. Contextual RL assumes the system can be parameterized by a set of contexts $\mathcal{C}$ and learns a function that maps each $c \in \mathcal{C}$ to an associated Markov decision process (MDP) \cite{ref:CRL1}. Recent variants exploit Bayesian methods for more sample-efficient training \cite{ref:CRL2}. Despite these advances, such methods still require retraining for any context not represented in $\mathcal{C}$. An emerging approach to improve robustness integrates RL with a stochastic model predictive control framework for nonlinear systems subject to unbounded process noise with closed-loop guarantees \cite{ref:MPC_Stochastic}. 

In contrast to DRL, the field of control theory has a long history of developing model-independent feedback control algorithms for unknown time-varying systems \cite{tsakalis1987adaptive,khalil_nonlinear_2002}. For systems of the form
\begin{equation}
\dot{x} = f(x,\theta(t),t) + g(x,t)u,
\end{equation}
adaptive controllers have been designed which can handle systems with time-varying parameters $\theta(t)$ and even systems in which the control gain $g(x,t)$ is time-varying, but has a known unchanging sign such that $|g(x,t)|>0, \forall t$ and the sign of $g$ is known. The challenging problem of stabilizing a system with unknown control direction, in which the sign of $g(x,t)$ is unknown, was solved by Nussbaum \cite{nussbaum1983some}, but maintained the requirement that the unknown sign cannot change with time i.e., $|g(x,t)|>0, \forall t$. This limitation was overcome by a recently developed approach of extremum seeking (ES) for the stabilization of unknown systems with unknown and time-varying control direction $g(x,t)$ which could pass through zero and change sign repeatedly \cite{scheinker2012minimum}.  

Bounded ES is a new form of ES which has a major advantage of guaranteed bounds on control efforts and parameter update rates despite acting on noisy and analytically unknown and time-varying systems \cite{scheinker2013model}. It has been studied for a wide range of dynamic systems \cite{scheinker2014extremum}, and a general weak-limit averaging analysis has extended its use with non-differentiable dithering functions as well as non-periodic time-varying systems \cite{scheinker2016bounded}. The ability to stabilize time-varying systems with guaranteed bounds on control effort has made bounded ES useful for safe hardware implementation on high energy systems, where abrupt parameter changes can easily result in damage, such as high energy charged particle accelerators \cite{scheinker2014hardware,scheinker2021extremum}. The bounded ES method has also been applied for GPS-denied source localization \cite{ghadiri2016new}, optimized path tracking in robotics \cite{bajpai2024investigating}, for biology-inspired 3D source seeking \cite{abdelgalil2022recursive}, and for tokamak stabilization \cite{de2022event}.

Recent research has combined RL with classical controllers to inject robustness, and safety into learning-based control. In \cite{romero2024actor}, a differentiable MPC is integrated with actor–critic policy, providing better real-time performance on agile robotics. In another work \cite{guha2021online}, MRAC-RL deploys a fast model-reference adaptive controller in the inner loop together with an outer-loop RL policy to mitigate simulation-to-reality mismatch. In \cite{emam2022safe}, robust control-barrier-function (CBF) layers project RL actions onto a certified safe set, improving safe exploration. These directions motivate a promising path toward hybrid controllers that integrate RL with traditional controllers.

While DRL is a powerful method for data-based learning of controllers for dynamic systems, one of the fundamental challenges is handling unknown and rapidly time-varying systems. On the other hand, while adaptive methods such as bounded ES are robust for time-varying systems, they are inherently local feedback-based schemes that do not exploit trajectory histories, are subject to suboptimal solutions, and can converge slowly in high-dimensional parameter spaces. This paper proposes a hybrid framework that combines DRL with bounded ES to leverage the strengths of both. We train a DRL policy using large datasets to find solutions in very few steps when the test dynamics are close to the training distribution. Crucially, the ES layer is warm-started from the DRL policy seeded with DRL-recommended control parameters, which reduces transients and accelerates adaptation when conditions drift. If the system starts to quickly change with time, and the learned DRL policy is no longer valid, robust bounded ES takes over and prevents the system's performance from severe degradation. 

We demonstrate our results with numerical studies of general time-varying dynamic systems, a detailed simulation study of a particle accelerator application with a time-varying magnetic lattice, mimicking natural accelerator behavior, and a detailed simulation study of an intermittent-contact robotic block-pushing task that pushes an object to a time-varying goal position.

\section{Background}

%
\subsection{Bounded Extremum Seeking for Time-Varying Systems}
\label{ES}
%

We briefly recall the results that summarize bounded ES as developed in \cite{scheinker2013model,scheinker2014extremum,scheinker2016bounded}. For $x\in\mathbb{R}^n$, consider system
\begin{equation}
	\dot{x} = f(x,t) + g(x,t)u(x,t), \label{dxdt}
\end{equation}
where $f:\mathbb{R}^{n}\times\mathbb{R} \rightarrow \mathbb{R}^n$ and $g:\mathbb{R}^{n}\times\mathbb{R} \rightarrow \mathbb{R}^{n\times m}$ are unknown, and $u: \mathbb{R}^{n}\times\mathbb{R} \rightarrow \mathbb{R}^m$ is the control. We focus on two special cases that are most relevant to this work. For stabilization, take $g:\mathbb{R}^{n}\times\mathbb{R} \rightarrow \mathbb{R}^{n}$ and a scalar controller $u:\in\mathbb{R}^{n}\times\mathbb{R} \rightarrow \mathbb{R}$. Choose a Lyapunov candidate $V(x)=x^Tx$, and define the feedback controller as
\begin{equation}
	u = \sqrt{\alpha\omega}\cos(\omega t + k V(x)), 
\end{equation}
then for large $\omega$, we can approximate $x(t)$ by the weak limit-averaged dynamics of $\overline{x}(t)$ given by
\begin{equation}
	\dot{\overline{x}} = f(\overline{x},t) -\frac{k\alpha}{2} g(\overline{x},t)g^T(\overline{x},t)\nabla_{\overline{x}}V(\overline{x}).
\end{equation}
Crucially, the averaged system's control direction is now a positive semidefinite matrix $g(\overline{x},t)g^T(\overline{x},t)\geq 0$, so we no longer have a control direction sign ambiguity, and we can stabilize the origin relative to $\overline{x}(t)$ by choosing sufficiently large gain $k\alpha>0$. Another special case of (\ref{dxdt}), which is most relevant for optimization, is when $f(x,t)=0$, $g(x,t)$ is a diagonal matrix with diagonal entries $g_i(x,t), i\in\{1,\dots,n\}$, $u$ is a vector, and we have access to measurements $y$ of a noise-corrupted and analytically unknown time-varying cost function $J(x,t)$ which we aim to minimize, so that our system takes on the form
\begin{equation}
	\dot{x}_i = g_i(x,t)u_i(x,t), \quad y=J(x,t)+n(t).
\end{equation}
For this setup, which is typical of optimization problems, we design our feedback controller as
\begin{equation}
	u_i = \sqrt{\alpha\omega_i}\cos(\omega_i t + k y(x,t)), \quad \omega_i = r_i\omega, \quad r_i \neq r_j, \label{uESopt}
\end{equation}
resulting in averaged system dynamics 
\begin{eqnarray}
	\dot{\overline{x}} &=& -\frac{k\alpha}{2} g(\overline{x},t)g^T(\overline{x},t)\nabla_{\overline{x}}J(\overline{x},t) \nonumber \\
	&& \Longrightarrow \dot{\overline{x}}_i = -\frac{k\alpha}{2} g^2_i(\overline{x},t)\frac{\partial J(\overline{x},t)}{\partial \overline{x}_i},
\end{eqnarray}
a gradient descent of the unknown $J(x,t)$. In both cases above, the proof in \cite{scheinker2016bounded} shows that for $x\in K$ for any compact set $K \subset \mathbb{R}^n$, for any $T>0$, and any desired $\epsilon>0$, there exists $\omega^*$ such that for all $\omega > \omega^*$ we can guarantee that 
\begin{equation}
	\| x(t) - \overline{x}(t) \| < \epsilon \quad \forall t\in [0,T],
\end{equation}
and this $T$ can be extended to infinity if $\overline{x}(t)$ converges to a stable equilibrium. Therefore, bounded ES is a powerful model-free tool for optimizing a time-varying analytically unknown noise-corrupted output function of a dynamic system or for stabilizing unknown time-varying systems. In what follows, we will use these properties of ES to bring robustness to DRL-based feedback controllers and high-dimensional optimizers.

%
\subsection{Deep Reinforcement Learning for Feedback Control}
%

We model DRL as a discounted Markov decision process $M=(S,A,p,r,\gamma)$ with $\gamma\in[0,1)$. At time $t$, the agent observes $s_t\!\in\!S$, applies a continuous control $a_t\!\in\!A$, receives $r_t=r(s_t,a_t)$, and transitions to $s_{t+1}\!\sim\!p(\cdot\,|\,s_t,a_t)$, with objective
\begin{equation}
J(\pi)=\mathbb{E}_\pi\!\Big[\sum_{t=0}^{\infty}\gamma^t r_t\Big].
\end{equation}
For high-dimensional continuous actions, deterministic actor–critic methods reduce variance by differentiating through a deterministic policy. The deterministic policy gradient (DPG) theorem \cite{silver2014deterministic} states that for a differentiable deterministic policy $\mu_\theta:S\!\to\!A$,
\begin{equation}
\label{eq:dpg}
\nabla_\theta J(\mu_\theta)=
\mathbb{E}_{s\sim \rho^{\mu_\theta}}
\!\left[\nabla_\theta \mu_\theta(s)\;\nabla_a Q^{\mu_\theta}(s,a)\big|_{a=\mu_\theta(s)}\right],
\end{equation}
where $\rho^{\mu_\theta}$ is the discounted state visitation distribution and
$Q^{\mu_\theta}(s,a)$ is the action–value function.

Deep Deterministic Policy Gradient (DDPG) instantiates~\eqref{eq:dpg} with deep function approximators and two stabilizing mechanisms from deep Q-learning: (i) an experience replay buffer for off-policy updates and (ii) slowly updated target networks for both actor and critic \cite{lillicrap2015continuous}. We maintain an actor $\mu_{\theta^\mu}$ and a critic $Q_{\theta^Q}$, with target networks $\mu_{\theta^{\mu'}}$ and $Q_{\theta^{Q'}}$ updated by Polyak averaging $\theta^{\mu'}\!\leftarrow\!\tau\theta^\mu+(1-\tau)\theta^{\mu'}$ and $\theta^{Q'}\!\leftarrow\!\tau\theta^{Q}+(1-\tau)\theta^{Q'}$, $0<\tau\ll 1$.
During data collection we use exploratory actions $a_t=\mu_{\theta^\mu}(s(t))+\varepsilon(t)$.
\paragraph{Critic update}
From a minibatch $\{(s_i,a_i,r_i,s'_i,d_i)\}_{i=1}^N$ sampled from replay $\mathcal D$, where $s'_i$ is the next state and $d_i\in\{0,1\}$ is the terminal indicator, the TD target and loss are
\begin{align}
y_i &= r_i + \gamma (1-d_i)\,
Q_{\theta^{Q'}}\!\big(s'_i,\,\mu_{\theta^{\mu'}}(s'_i)\big),\\[-1mm]
\mathcal{L}(\theta^Q) &= \tfrac{1}{N}\sum_{i=1}^N
\big(Q_{\theta^Q}(s_i,a_i)-y_i\big)^2 .
\end{align}

\paragraph{Actor update}
Approximating the DPG with samples from replay $\mathcal D$,\begin{equation}
\label{eq:actor-grad}
\nabla_{\theta^\mu} J \;\approx\;
\tfrac{1}{N}\sum_{i=1}^N
\Big[\nabla_a Q_{\theta^Q}(s,a)\big|_{a=\mu_{\theta^\mu}(s_i)}\Big]\;
\nabla_{\theta^\mu}\mu_{\theta^\mu}(s_i).
\end{equation}

%
%
%
\subsection{Motivating Example of a Time Varying System}

\begin{figure}
  \centering
  \subfloat[Low $f$\label{fig:rl_es_lowfreq}]
  {\includegraphics[width=0.48\linewidth]{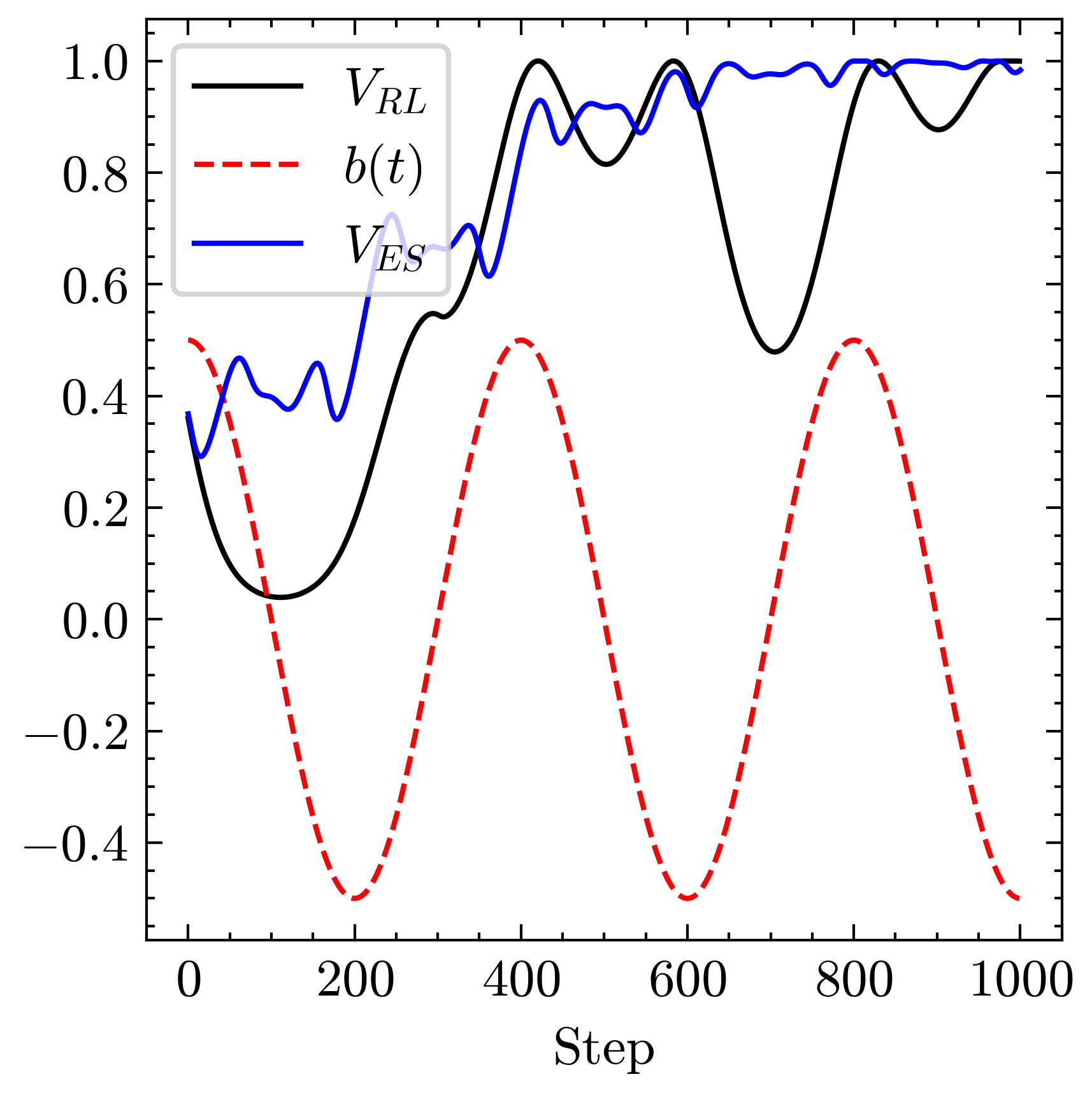}}
  \hfill
  \subfloat[High $f$\label{fig:rl_es_highfreq}]
  {\includegraphics[width=0.48\linewidth]{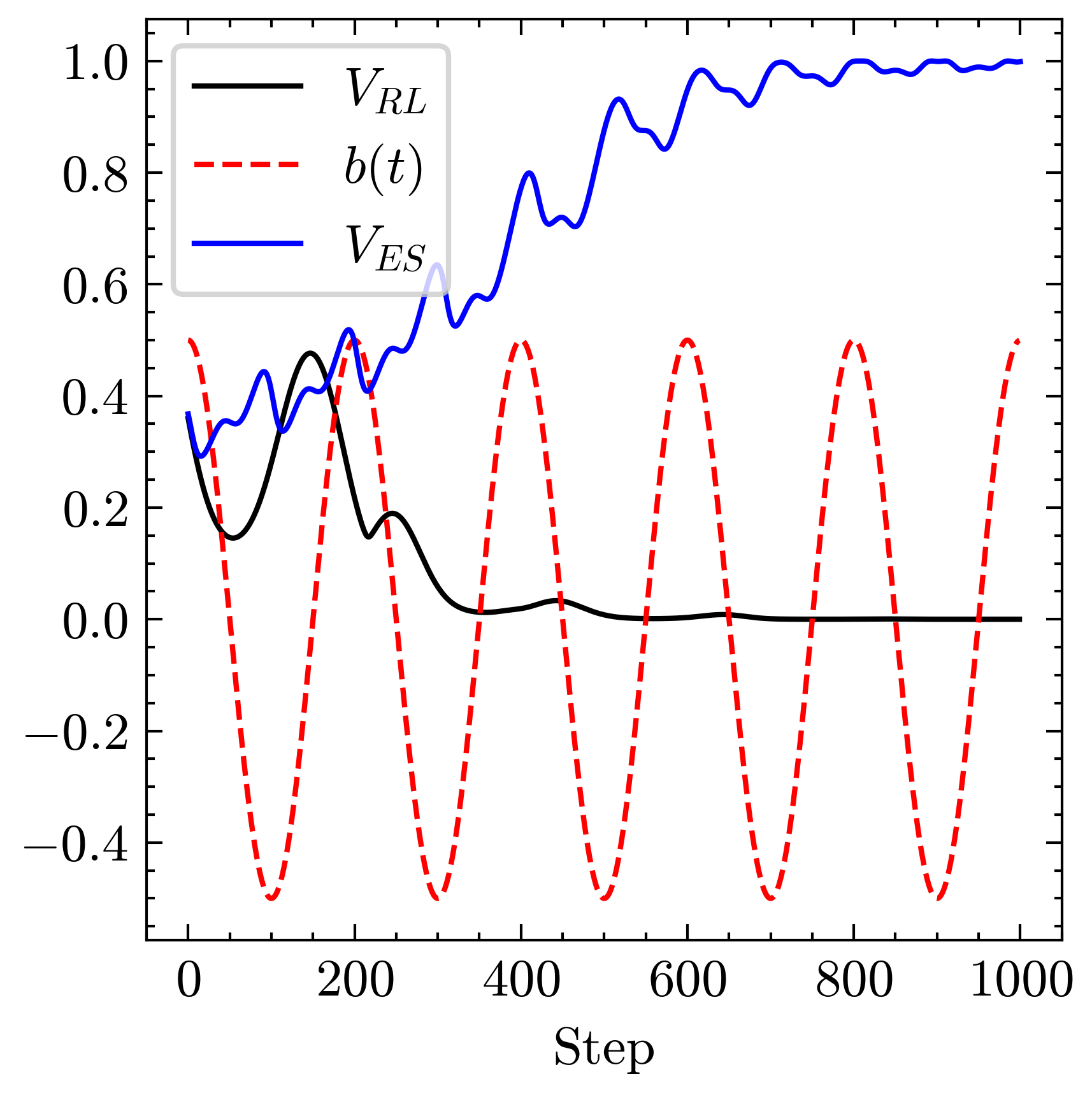}}
  \caption{Maximizing $V(x)=\exp(-x^{2})$ under a sinusoidally varying control direction $b(t)=b_0\cos(2\pi f t)$. 
  (a) Low $f$: DRL reaches high $V$ temporarily, but diverges during large swings of $b(t)$; ES reaches and maintains high $V$. 
  (b) High $f$: DRL diverges; ES maintains high $V$ after convergence.}
  \label{fig:rl_es_freq}
\end{figure}

We consider a simple 1D open-loop unstable linear time-varying system with unknown time-varying control direction
\begin{equation}
	\dot{x} = ax + b(t)u, \quad b(t) = b_{0}\cos(2\pi f t), \quad V(x)=e^{-x^2},
\end{equation}
where $a, b_0>0$ and $f$ are unknown and our goal is to maximize the analytically unknown objective function $V(x)$ based only on samples. Bounded ES makes $x=0$ a practically stable equilibrium of this system by 
\begin{equation}
    u=\sqrt{\alpha\omega}\cos(\omega t - kV), 
\end{equation}
which for large $\omega$ gives averaged dynamics
\begin{equation}
	\dot{\overline{x}} = a\overline{x} + \frac{k\alpha}{2} b^2(t) \nabla V(\overline{x}).
\end{equation}
The averaged dynamics control direction is no longer unknown because $b^2\geq0$ and therefore sufficiently large $k\alpha > 0$ ensures a gradient ascent of $V(\overline{x})$. As detailed in the bounded ES references, $\omega$ must be chosen sufficiently large relative to $f$. While trivial for bounded ES, this problem is extremely difficult for DRL as shown in Fig. \ref{fig:rl_es_freq}. 

When $b(t)$ varies slowly (low $f$; Fig. \ref{fig:rl_es_lowfreq}), ES successfully maximizes the objective $V(x)$ by driving $x\!\to\!0$. In contrast, the RL policy can reach the maximizer but fails to remain there as $b(t)$ continues to drift. As the drift rate increases (Fig. \ref{fig:rl_es_highfreq}), the plant moves out of the RL policy’s training distribution and the attained $V$ degrades. In contrast, ES continues to ascend $V$ due to its averaging-based gradient property even as the control direction flips.

\section{ES-DRL for Particle Accelerators}

Large particle accelerators are intrinsically time-varying; they are composed of thousands of coupled electromagnetic components whose characteristics drift with temperature, usage, and maintenance; diagnostics are limited and noisy, and they require continual manual retuning. The amplitude and phase of the resonant radio frequency (RF) accelerating cavities  drift with time due to temperature, amplifier/distribution effects, and magnet power-supplies ripple and hysteresis break magnetic–field repeatability. Such time-varying behavior is documented for the kilometer-long Los Alamos Neutron Science Center (LANSCE) linear accelerator, for which adaptive ES methods have been developed for beam loss minimization \cite{scheinker2020extremum}. Because ES is a robust local search-based technique, as the number of parameters increases, convergence can take longer and can get stuck in local minima. Therefore, we are studying the application of a combined ES-DRL controller that can learn and quickly approximately optimize such systems with large numbers of parameters, and then robust ES can keep things stable even as they drift with time.

\subsection{Simulator}

We model the initial 12-meter long section of LANSCE, the low energy beam transport (LEBT) using the Kapchinskij–Vladimirskij (KV) envelope model. The KV equations were initially introduced in 1959 \cite{kapchinskij1959limitations}; they model space-charge effects in a uniform elliptical beam moving through linear focusing fields. They capture the dominant transverse beam physics while remaining computationally efficient for learning-in-the-loop experiments \cite{lund2004stability}.

The KV equations are two second-order ODEs, where for a \textbf{fixed time} $t$, the ODEs are solved over the independent variable $z$, where $z$ represents the distance along the beamline in the direction of the beam travel.

\begin{figure*}[t] 
    \centering
    \includegraphics[width=0.7\textwidth]{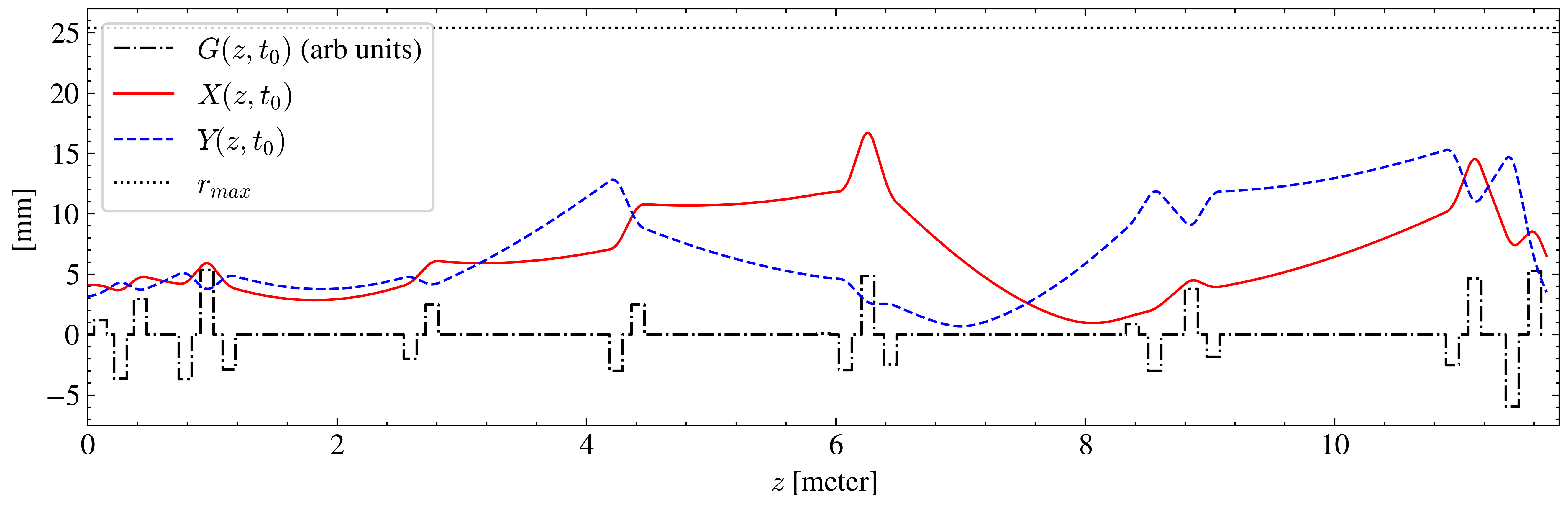}
    \caption{Solutions of the KV equations based on the 22 quadrupole magnet strengths $(G(z, t_0))$ at initial time $t_0$.}
    \label{fig:Beam_KV}
    \vspace{-12pt}
\end{figure*}

Let $X(z, t)$ and $Y(z, t)$ denote the rms-like transverse envelope radii in the horizontal and vertical planes. The quadrupole magnet lattice enters through a signed focusing profile $G(z; u(t))$ (units of T/m) that flips sign with magnet polarity. With normalized emittances $\varepsilon_x,\varepsilon_y$ and generalized perveance $K$ (proportional to beam current and inversely to $\gamma^3\beta^3$), the KV envelope dynamics are written as
\begin{align}
X''(z, t) &= -\,\kappa(z, t)\,X(z, t) \;+\; \frac{\varepsilon_x^2}{X^3(z ,t)}
\;+\; \frac{K}{X(z, t)+Y(z, t)}, \label{eq:kvx}\\
Y''(z, t) &= \phantom{-}\kappa(z,t)\,Y(z, t) \;+\; \frac{\varepsilon_y^2}{Y^3(z, t)}
\;+\; \frac{K}{X(z, t)+Y(z, t)}, \label{eq:kvy}
\end{align}
where $\kappa(z, t)\coloneq G(z; u(t))/(\beta\rho)$ and $X'',Y''$ represent $\tfrac{d^2 X}{d z^2},\tfrac{d^2 Y}{d z^2}$. The $\kappa$ terms encode linear focusing/defocusing and the $K$ term models space–charge defocusing. An example solution of the KV equations for the LANSCE LEBT at time $t = t_0$ is shown in Fig. \ref{fig:Beam_KV}. The LANSCE LEBT contains $N=22$ independently driven quadrupole magnets. We map the controller's action vector $u\in\mathbb{R}^{22}$ (one element per
magnet) to a piecewise-constant profile as
\begin{equation}
G(z; u(t)) \;=\; \sum_{i=1}^{22} u_i(t)\, b_i(z), \qquad
b_i(z)=\sigma_i\,\mathbf{1}_{[z_i,\;z_i+\ell_i]}(z),
\label{eq:G-map}
\end{equation}
where $z_i$ and $\ell_i$ are the longitudinal location and effective length of magnet $i$, where $i \in (1,...,22)$, $\sigma_i\!\in\!\{+1,-1\}$ encodes polarity, and $\mathbf{1}$ is an indicator function for the magnet gap. Drift sections (where the function $G(z; u(t)) = 0$) contain no magnets and in these sections the beam is not subject to external fields. See \cite{williams2024safe} for additional simulation details on implementing the KV equations. 

We integrate \eqref{eq:kvx}--\eqref{eq:kvy} forward in $z$ with fixed-step fourth-order Runge-Kutta. Initial conditions $\big(X(0,t),Y(0,t),X'(0,t),Y'(0,t)\big)$ represent the incoming beam at the LEBT entrance at time $t$. For initial experiments these values are kept fixed, and are randomized later to improve robustness of the training process.

The four-dimensional observation, which is provided as the state to the RL controller and used to calculate the cost in the ES controller at plane $z$ at time $t$ is defined as:
\begin{equation}
o(t) \;=\; \big[X(z, t),\; Y(z, t),\;
X'(z, t),\; Y'(z, t)\big],
\label{eq:obs}
\end{equation}
where $o(t)$ contains the full transverse profile of the beam for all \(z\in[0,z_{\max}]\).

The KV model balances fidelity and speed while accurately describing the physics of a space–charge dominated, nonrelativistic, and high-current beam. 

\begin{figure*}[t]
  \centering
  \includegraphics[width=0.98\linewidth]{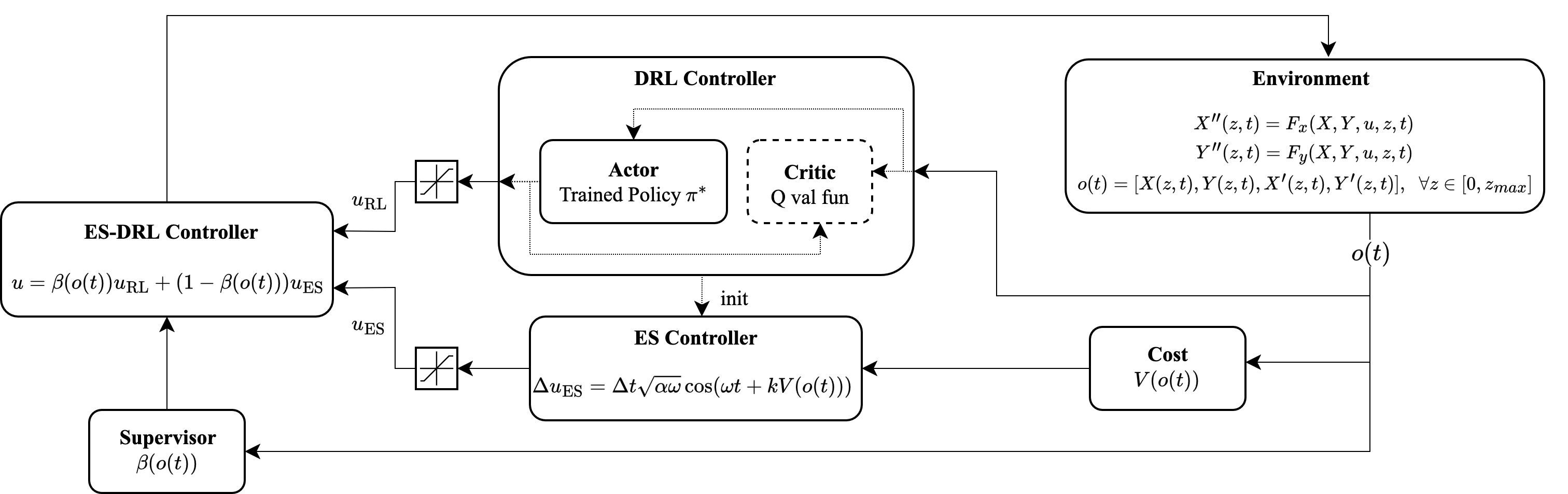}
  \caption{Architecture of the ES--DRL controller for accelerator tuning. 
A supervisor selects binary $\beta \in \{0,1\}$  based on safety constraints and combines 
$u = \beta(o(t))u_{\mathrm{RL}} + (1-\beta(o(t)))u_{\mathrm{ES}}$. 
ES may be warm-started from DRL (dotted).}
  \label{fig:ESDRL_controller}
  \vspace{-12pt}
\end{figure*}
\subsection{DRL Training Procedure}
\label{DRL_training}
\begin{table}[t]
\centering
\caption{DDPG training hyperparameters}
\begin{tabular}{lll}
\toprule
\textbf{Hyperparameter} & \textbf{Particle Accelerator} & \textbf{Robot}\\ 
\midrule
State $(s)$ dimension & $4\times 4000$ & 28\\
Action $(a)$ dimension & $22$ & 4\\
Discount $\gamma$ & $0.99$ & 0.99\\
Actor learning rate & $1\times 10^{-5}$  &$1\times 10^{-4}$ \\
Critic learning rate & $1\times 10^{-4}$ &$1\times 10^{-4}$\\
Replay buffer $|\mathcal{D}|$ & $10^6$ transitions (uniform) &$10^6$ transitions\\
Batch size $B$ & $128$ &256\\
Target update $\tau$ & $0.005$ (soft, every step) &0.005\\
Exploration noise & $\mathcal{N}(0, 0.1)$ per action &$\mathcal{N}(0, 0.1)$ \\
\bottomrule
\end{tabular}
\label{DDPG_params}
\end{table}

The state stacks KV beamline envelopes $s=\big[X,\,Y,\,X',\,Y'\big]$ over $N{=}4000$ longitudinal locations, yielding $s\in\mathbb{R}^{16000}$. Actions are continuous setpoints for $22$ quadrupole magnets, $a\in\mathbb{R}^{22}$, applied within machine-valid limits.

\emph{Reward Structure:}
We adopt a measurement-aligned reward that increases when the beam stays compact along the line, is well aligned at the end station, and varies smoothly. Let $z\in[0,z_{\max}]$ denote the longitudinal coordinate, with transverse envelopes $X(z),Y(z)$ and slopes $X'(z),Y'(z)$. 
Define the hinge $\langle a\rangle_{+}\triangleq\max(0,a)$. 
We then form the averaged path envelopes
\begin{equation}
\overline X(t)=\frac{1}{N}\sum_{k\in Z} X(k,t),\qquad
\overline Y(t)=\frac{1}{N}\sum_{k\in Z} Y(k,t),
\label{eq:avg}
\end{equation}
where \(Z=\{0,\Delta z,2\Delta z,\dots,z_{\max}\}\), \(\Delta z=2.92  \,\si{mm}\), and \(N=4000\) is the number of grid points.

We take $r_{\max}=25.4  \si{mm}$ as an operational radius bound (Fig.~\ref{fig:Beam_KV}) and set the envelope band and terminal target as
\begin{equation}
  r_{\text{band}} \;=\; \tfrac{1}{2}r_{\max}, 
  \qquad
  r_{tt}^2 \;=\; \tfrac{1}{2}r_{\max}^2.
\end{equation}

\paragraph*{Penalty composition}
The cumulative penalty is the sum of envelope, smoothness, and terminal terms:
\begin{equation}
  P \;=\; P_{\text{env}} + P_{\text{smooth}} + P_{\text{term}},
\end{equation}
\begin{equation}
  P_{\text{env}} \;=\; w_e\Big(\,\big\langle\,\overline{X}-r_{\text{band}}\,\big\rangle_{+}
                             \;+\;\big\langle\,\overline{Y}-r_{\text{band}}\,\big\rangle_{+}\Big),
\end{equation}
\begin{equation}
  P_{\text{smooth}} \;=\; w_s\Big(\overline{X'^2}+\overline{Y'^2}\Big).
\end{equation}

The averaged values of $\overline{X'^2}$ and $\overline{Y'^2}$ are computed similarly to those given in Eq. \eqref{eq:avg}.

At $z=z_{\max}$, the terminal cost drives the beam to become circular (equal $X$ and $Y$), flat (small derivatives), and of a prescribed radius $r_{tt}$ according to:
\begin{equation}
\begin{aligned}
P_{\text{term}} &= w_r\,\bigl|X(z_{\max}, \cdot)-Y(z_{\max}, \cdot)\bigr|
                  + \\
                &\quad w_w\bigl(|X'(z_{\max}, \cdot)| +  |Y'(z_{\max}, \cdot)|\bigr)  + \\
                &\quad w_t\,\bigl|X(z_{\max}, \cdot)^{2}+Y(z_{\max}, \cdot)^{2}-r_{tt}^{2}\bigr|.
\end{aligned}
\end{equation}
$w_e,w_s,w_r,w_w,w_t$ denote the weights.

The instantaneous reward is the bounded inverse
\begin{equation}
  R \;=\; \frac{1}{1+P}\in(0,1],
\end{equation}
which increases monotonically as envelope excursions, slope, and terminal mismatches decrease.

If the IVP solver fails to return valid envelopes we assign a large penalty and terminate the episode. The environment is then reset to the last feasible setting to continue exploration from a known-good configuration near the constraint boundary rather than repeatedly returning to the initial operating point. Directly training a controller over all $22$ quadrupole inputs caused frequent failures of the initial–value problem (IVP) solver for certain magnet combinations, which prevented the simulator from returning a solution to the KV equations. To stabilize learning while preserving the intended control objective, we adopt a curricular procedure that progressively increases the control dimension.

\emph{Phase I-Group-wise training (stabilization)}:
We partition the $22$ quadrupoles into seven longitudinally contiguous groups. One group is trained at a time while the rest are held at nominal settings. For group $g$, the actor produces incremental updates, the IVP for \eqref{eq:kvx}–\eqref{eq:kvy} is solved, and the reward is evaluated. Groups are visited sequentially, carry forward the actor--critic parameters to preserve lattice knowledge and stabilize learning.

\emph{Phase II-Full $22$-input training (coordination)}:
After the group-wise pass, all $22$ quadrupole magnets are trained together and training continues with the actor–critic initialized from Phase~I. We use fixed initial beam conditions $(X(0, \cdot),Y(0, \cdot),X'(0, \cdot),Y'(0, \cdot))$ to promote coordinated moves and prevent solver failures, aligning the controller with the final deployment objective in the full actuation space.

\emph{Phase III-Robust training (Random initial conditions)}:
To improve robustness, we randomize the initial beam state each episode drawing \(X(0, \cdot),Y(0, \cdot)\sim\mathcal{U}[1.5\times10^{-3} m,\,4.5\times10^{-3} m]\) and \(X'(0, \cdot),Y'(0, \cdot)\sim\mathcal{U}[-10^{-2},\,10^{-2}]\) and continue full 22 input training.

Curriculum transitions (Phase~I$\to$II$\to$III) are triggered by the reward saturation test and a minimum-episode budget per phase.
Across all phases we use an off-policy deterministic actor-critic (DDPG) with experience replay, slowly updated target networks \cite{lillicrap2015continuous}. We adopt the settings in Table~\ref{DDPG_params}. The actor $\mu_\theta$ is a 3-layer MLP with 512 units per layer, LayerNorm and ReLU nonlinearities, followed by a $\tanh$ output that scales actions to the permitted range. The critic Q value function is a 3-layer MLP (512 units per layer, LayerNorm+ReLU) applied to the concatenated $(s,a)$ and outputs a scalar value. 

\emph{Runtime policy:}
We deploy the policy trained by DDPG. At run time we use only the frozen actor $\mu_\theta$ to prevent the policy drift and define the RL control command as
\begin{equation}
u_{\mathrm{RL}}(o(t)) \;=\; \operatorname{sat}\!\big(\mu_\theta(o(t))\big),
\end{equation}
where $o(t)$ is the current observation and $\operatorname{sat}(\cdot)$ enforces elementwise actuator limits. Exploration noise is disabled at evaluation. The critic and target networks are used only during training and are not invoked online.

\subsection{Combined ES-DRL Controller}
%
We consider time-varying accelerator tuning in which the control input consists of the 22 quadrupole magnet strengths $Q=(Q_1,\dots,Q_{22})$. In Fig.~\ref{fig:Beam_KV}, each \(Q_i\) corresponds to one positive or negative peak of the gradient profile \(G(z)\). We propose the combined ES--DRL iterative controller
\begin{equation}
    Q_i(t+1) = u(o(t),t,Q(t), Q(0)),
\end{equation}
\begin{equation}
  u(o(t),t)=\beta(o(t))\,u_{\mathrm{RL}}(o(t))
  +\bigl(1-\beta(o(t))\bigr)\,u_{\mathrm{ES}}(o(t),t),
  \label{eq:ESRL}
\end{equation}
whose architecture is shown in Fig.~\ref{fig:ESDRL_controller}. We let \(o(t)=\bigl[X(z,t),\,Y(z,t),\,X'(z,t),\,Y'(z,t)\bigr]\) denote the observation vector, where each component is the corresponding trajectory sampled along \(z\in[0,z_{\max}]\) on a uniform grid of 4000 points, with \(z_{\max}=11.70\) m. In this setup, the RL-based controller always updates magnet settings $Q_i$ relative to their initial conditions according to:
\begin{equation}
    Q_i(t+1) = \underbrace{Q_i(0) + \hat{u}_{RL,i}(o(t))}_{u_{RL,i}}.
\end{equation}
The ES controller instead uses a finite difference approximation of Eq. \eqref{uESopt}:
\begin{equation}
    \dot{Q}_i(t) \approx \frac{Q_i(t+1)-Q_i(t)}{\Delta_t}=\sqrt{\alpha\omega_i}\cos(\omega_i \Delta_t t - k V(o(t))),
\end{equation}
where $\Delta_t \ll 1$ is sufficiently small relative to $\max{\omega_i}$ for the approximation to hold, which gives
\begin{equation}
Q_{i}(t+1)
= \underbrace{Q_{i}(t)
  + \Delta t\,\sqrt{\alpha \omega_i}\,
    \cos\!\bigl(\omega_i t \Delta t - k\,V(o(t))\bigr)}_{u_{ES,i}}.
    \label{eq:ES}
\end{equation}

\emph{DRL controller:}
We train a Deep Deterministic Policy Gradient (DDPG) agent. At evaluation, only the trained actor \(\pi^\star\) is used. It maps observations to actions that are subsequently passed through a saturation block to enforce actuator limits. The critic (action–value) function \(Q_{\mathrm{val}}\) is used exclusively during training; see Section~\ref{DRL_training} for details.

\emph{ES controller:}
To reduce transients and accelerate adaptation, the ES controller is warm-started with the RL actor’s output. The ES objective is to maximize the same reward \(V(o(t))\). We use \(\alpha>0\) and distinct \(\omega_i>0\) as dither parameters, and  set feedback gain \(k = 15\).

\emph{Safety supervisor:}
The supervisor generates the binary switch \(\beta(o(t))\) from envelope measurements. It is evaluated at every discrete control step, so the handoff occurs immediately once the envelope threshold is violated. Beam loss increases as the beam envelope approaches the beam–pipe aperture; we enforce a safety margin by constraining the beam to remain within \(70\%\) of \(r_{\max}\).
Given the RL action \(u_{\mathrm{RL}}\), we integrate the KV equations \eqref{eq:kvx}–\eqref{eq:kvy} to obtain \(X(z,t)\) and \(Y(z,t)\) along the line (cf. Fig.~\ref{fig:Beam_KV}).

Let \(r_{\max}\) denote the allowable beam–pipe radius (Fig.~\ref{fig:Beam_KV}), with \(r_{\max}=25.4  \si{mm}\).
The switching law is
\begin{equation}
\beta \coloneq
\begin{cases}
1, & \text{if } \overline{X}(t)\ \text{or}\ \overline{Y}(t) < 0.7\,r_{\max} \text{(RL mode)},\\[2mm]
0, & \text{otherwise (ES mode),}
\end{cases}
\label{eq:beta}
\end{equation}
so that RL is used when the average envelopes are well within the aperture and control reverts to the robust ES component when either envelope approaches the aperture, which indicates higher beam loss.
Other beam-loss models may also be used \cite{williams2024safe}.
Applying this rule in Eq.\eqref{eq:ESRL} switches to ES whenever the RL policy violates the constraint or fails to stabilize the plant. The ES weak-limit averaging results hold and the overall ES-DRL magnet field dynamics evolve according to
\begin{equation}
  \dot{Q}(t) \approx \frac{k\alpha}{2}\nabla_{Q}V(o(t)),
  \label{eq:Q_dot}
\end{equation}
a local model-independent gradient ascent of the time-varying $V(o(t))$, relative to the controlled subset of $Q$. 

\subsection{Simulation Results}
\label{sec:robustness-eval}

\begin{figure}
  \centering
  \includegraphics[width=0.98\linewidth]{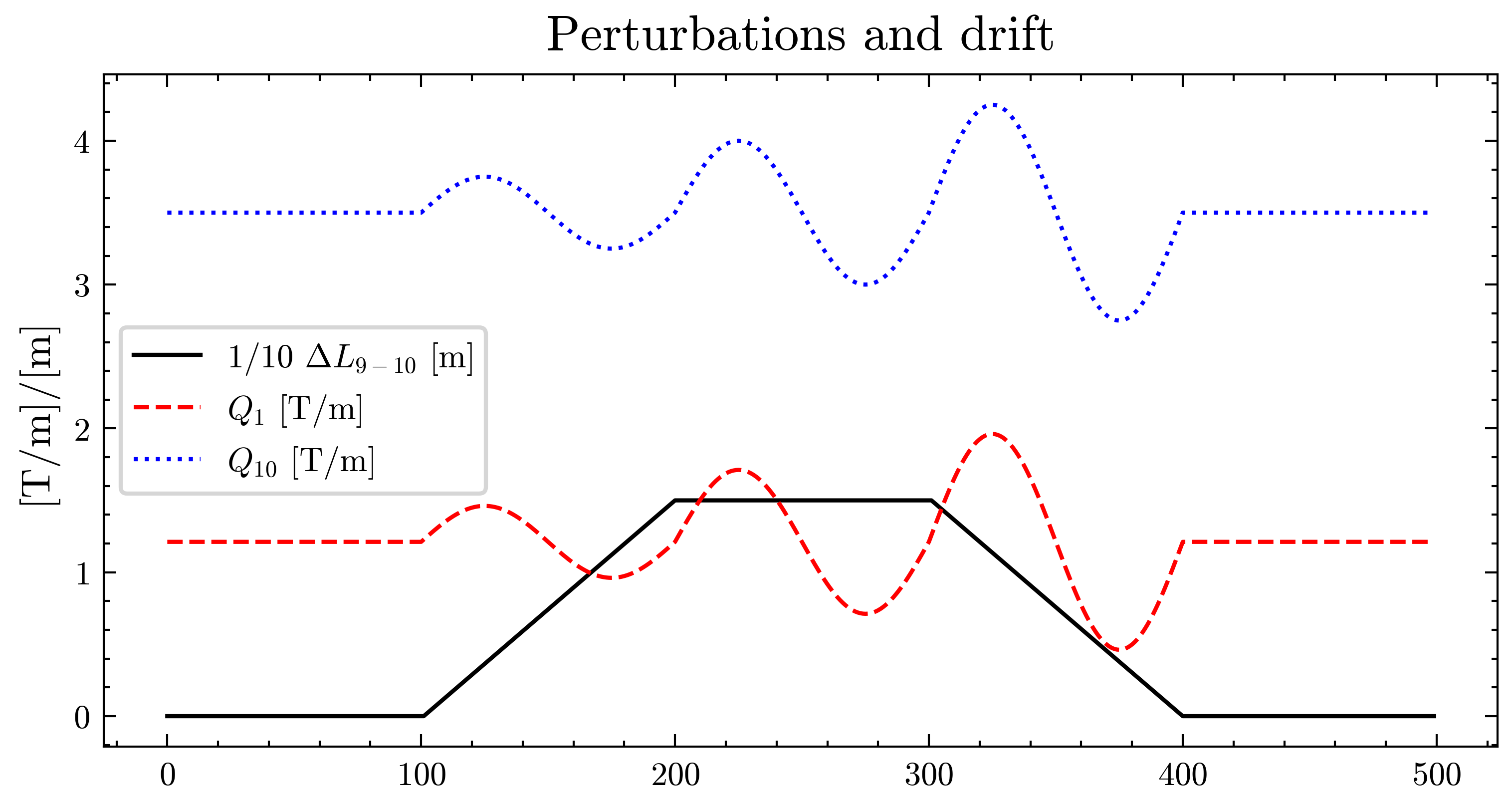}
  \includegraphics[width=0.99\linewidth]{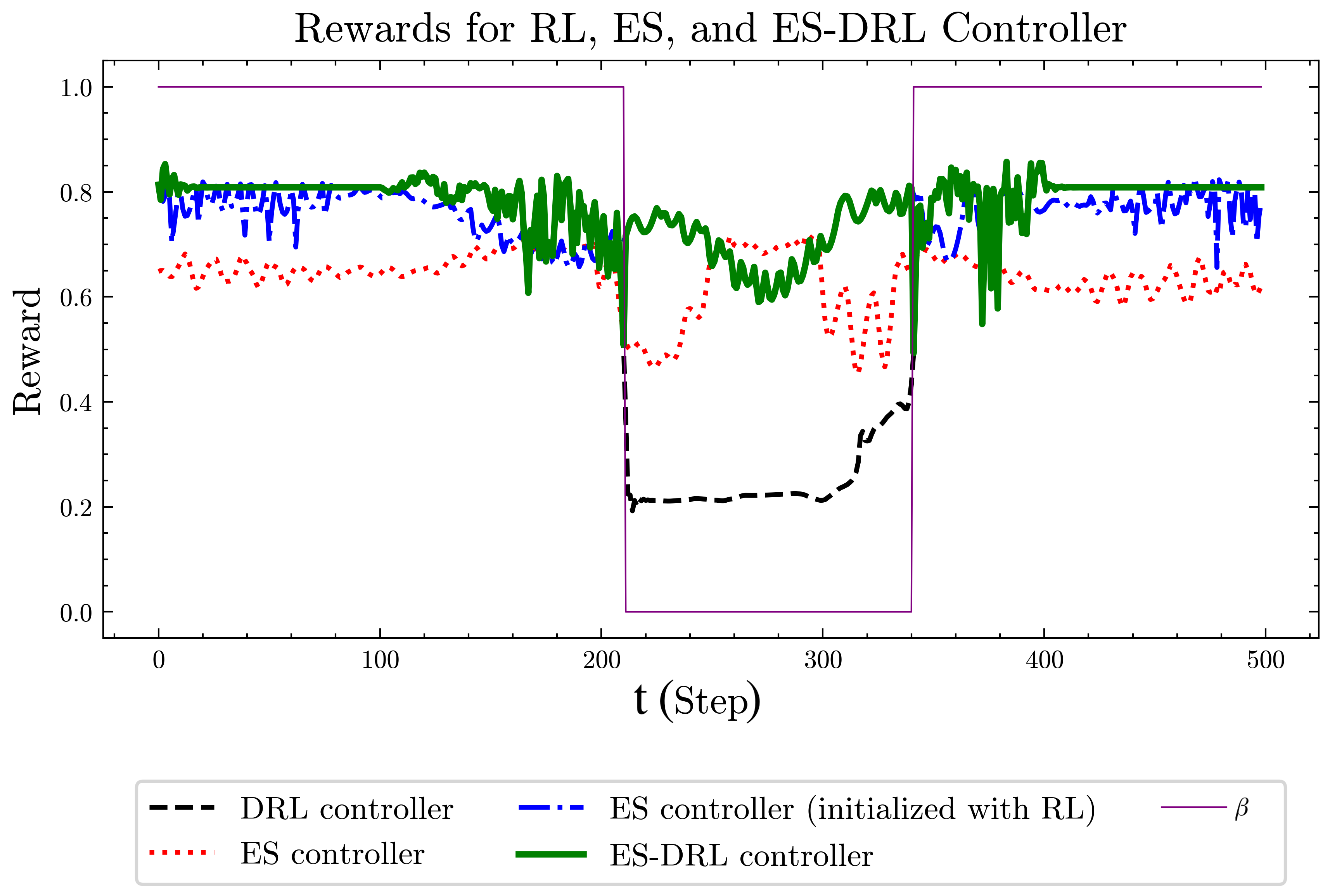}
  \caption{Perturbations and performance: (top) injected sinusoidal perturbations at \(Q_1\) and \(Q_{10}\) and the drift in segment length between \(Q_{9}\) and \(Q_{10}\) over 500 steps. (bottom) Resulting reward trajectories; the hybrid ES--DRL controller achieves the best overall reward.}
  \label{fig:combined}
  \vspace{-12pt}
\end{figure}

We evaluate four controllers in the KV-based simulator: 
(i) a DDPG policy (\textsc{DRL}); 
(ii) bounded extremum seeking (\textsc{ES}); 
(iii) \textsc{ES} warm-started with the \textsc{DRL} action at engagement; and 
(iv) the proposed combined \textsc{ES--DRL} controller.

In this experiment the agent’s action space \emph{excludes} \(Q_1\) and \(Q_{10}\). These two magnets are driven exogenously by sinusoids, and \(Q_{10}\) is additionally perturbed via a geometric drift. To probe generalization beyond training, we apply a perturbation schedule that pushes the policy far outside its training distribution. Specifically, we excite \(Q_1\) and \(Q_{10}\) with discrete-time, amplitude-ramped sinusoids
\[
Q_j(t) \;=\; Q_j^\star + A(t)\,\sin\!\bigl(\nu t\bigr), 
\qquad j\in\{1,10\},\; t=0,\dots,500,
\]
with setpoints \(Q_1^\star = 1.21\si{T/m}\) and \(Q_{10}^\star = 3.5 \si{T/m}\).
The angular frequency is \(\nu=\pi/50\) rad/step (period \(100\) steps).
At step 100, the amplitude starts to increase from 0.25 to 0.75 over 400 steps and then drops back to 0, as shown in the top part of Fig.~\ref{fig:combined}.
Here \(Q_1\) and \(Q_{10}\) correspond to the first and tenth extrema of \(G(z)\) in Fig.~\ref{fig:Beam_KV}.

We also introduce a \emph{geometric drift} by shifting the longitudinal location of \(Q_{10}\) in \eqref{eq:kvx}--\eqref{eq:kvy}:
\[
L_{9-10}(t) \;=\; L_{9-10}^\star + \Delta L_{9\text{--}10}(t),
\]
where \(L_{9-10}^\star = 176\,\si{mm}\) is the distance between $Q_9$ and $Q_{10}$. The inter-magnet spacing \(\Delta L_{9\text{--}10}(t)\) is ramped from \(0\) to \(150\,\si{mm}\) over steps \(t=100{:}200\), held at \(0.15\,\si{m}\) for \(t=200{:}300\), and ramped back to \(0\) over \(t=300{:}400\), as shown in the top part of Fig.~\ref{fig:combined}. Aggregate performance under these perturbations is summarized in Fig.~\ref{fig:combined}.

Because \(Q_1\) and \(Q_{10}\) are not actuated by the controller, the agent must compensate by coordinating the \emph{remaining} 20 quadrupoles. Note that the \textsc{DRL} policy was trained to command all 22 magnets; in this test it must recover performance without authority over \(Q_1\) and \(Q_{10}\).

\emph{Findings:} 
The standalone \textsc{DRL} policy maintains a high reward (\(\approx 0.8\)) up to \(\sim\)step \(160\), when \(\Delta L_{9\text{--}10}\!\approx\!100\,\si{mm}\) and the sinusoid amplitude is \(0.25\).
As the amplitude increases and the spacing reaches \(150\,\si{mm}\), the \textsc{DRL} reward degrades (out-of-distribution behavior), the supervisor control hands off to \textsc{ES}, keeping updates bounded.
The combined \textsc{ES-DRL} controller maintains rewards well above \(0.6\) throughout the plateau and the return ramp.
When the spacing decreases toward \(100\,\si{mm}\) and the perturbation weakens, the \textsc{DRL} policy recovers and \(\beta\) returns toward \(1\), restoring fast, coordinated adjustments.
Warm-starting \textsc{ES} with \textsc{DRL} actions improves transients relative to standalone \textsc{ES}.
Overall, the hybrid ES-DRL controller achieves the highest and most stable reward trajectory over the full 500 steps, consistent with Fig.~\ref{fig:combined}.


\section{ES-DRL controller for Intermittent Contact robot task with Time-Varying Goal}
\label{sec:robotics}
\begin{figure}
  \centering
  \includegraphics[width=0.975\linewidth]{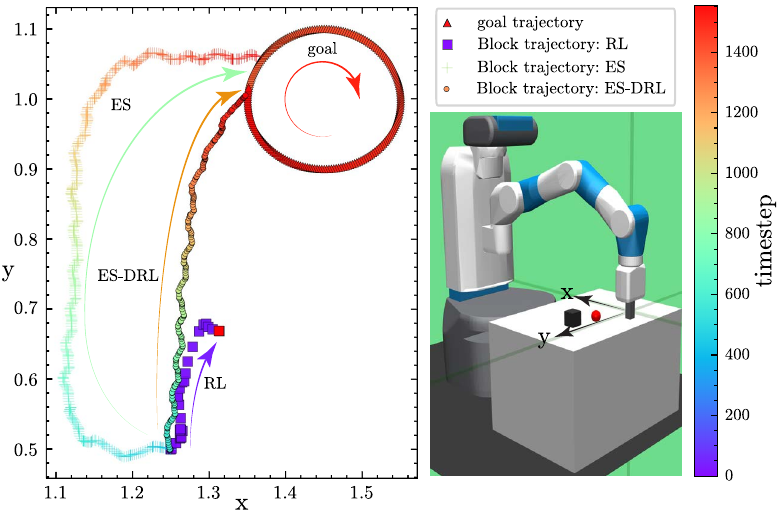}
  \caption{Time-varying goal trajectory and block position. }
  
  \label{fig:robot}
  \vspace{-12pt}
\end{figure}

Next, we apply the same ES-DRL controller in \eqref{eq:ESRL} to the control of a robotic arm, as shown in Fig. \ref{fig:robot}. The task consists of a 7-DoF Fetch mobile manipulator arm that must push a movable block on a tabletop to a desired goal position in the FetchPush benchmark environment \cite{plappert2018multi}. Robotic block pushing is a widely studied manipulation task, with recent work demonstrating robust pushing using force feedback alone \cite{heins2024force}. The desired goal follows a smooth time-varying trajectory in the tabletop plane: at each step $t$ it traces a circular path of radius $0.10$~m about a nominal center $(g_{x0},g_{y0})$ with a period of $200$ steps, i.e., $g_x(t)=g_{x0}+0.10\sin(2\pi t/200)$ and $g_y(t)=g_{y0}+0.10\cos(2\pi t/200)$, and $g_z(t)$ is fixed, which produces a distribution shift relative to the stationary-goal settings used for RL training. The observation is a 25-dimensional vector containing the end-effector Cartesian position and velocity, the block position, the relative block-to-gripper position, the left and right gripper finger joint displacements and velocities, the block XYZ Euler orientation, and the block linear and angular velocities. In our implementation we concatenate observation and desired\_ goal to form a state vector $s_t \in \mathbb{R}^{28}$. Actions are continuous Cartesian displacement commands for the end-effector together with a gripper command $g(t)$,
\begin{equation}
a_t \in \mathbb{R}^{4}, \qquad a_t = [\Delta x(t),\, \Delta y(t),\, \Delta z(t),\, g(t)]^\top,
\end{equation}
which are applied through the MuJoCo simulator’s internal mocap operational-space controller to generate the low-level joint actuation. 
The DDPG training hyperparameters are summarized in Table \ref{DDPG_params}. During evaluation, exploration noise is disabled and only the trained actor is used to compute the RL actions.

\emph{Reward design:} 
We use a simple dense shaping reward based on two Euclidean distances: $d_1$, between the end-effector and the block, and $d_2$, between the block and the goal. At each step we set $r_t = -(d_1+d_2)$, and add a success bonus of $+2$ when the block reaches the goal.

\emph{Safety supervisor:} 
For ES-DRL controller, we use RL for fast approach to the goal and switch to bounded ES once physical interaction begins. Let $c_t\in\{0,1\}$ be a contact flag that indicates whether the end-effector is in contact with the block at time $t$. We define a supervisor
\begin{equation}
\beta_t \in \{0,1\},\qquad
\beta_t =
\begin{cases}
1, & c_t = 0 \quad (\text{RL mode}),\\
0, & c_t = 1 \quad (\text{ES mode}).
\end{cases}
\label{eq:fetch_switch}
\end{equation}

\emph{Results:}
Fig.~\ref{fig:robot} highlights the effect of goal drift as an out-of-distribution disturbance for the learned policy. Although the RL controller initially approaches and begins to push the block, the time-varying target drives the policy outside the training distribution; the gripper subsequently loses effective contact and the block stagnates away from the moving goal. ES is more robust to this nonstationarity and eventually discovers a pushing direction, but it requires a longer, more exploratory approach to first acquire contact and then align the push, resulting in a larger path length. In contrast, ES-DRL leverages RL for rapid and directed approach to establish contact, then switches to ES during interaction to adapt the push online, preserving contact and reaching the time-varying goal more quickly and with a more direct trajectory.

\section{Conclusions}
We presented a hybrid ES–DRL control framework that leverages the complementary strengths of deep reinforcement learning and bounded extremum seeking for stabilizing and optimizing nonlinear time-varying systems. Numerical studies of general time-varying systems, particle accelerator beams, and robot arms demonstrated that while DRL policies can achieve rapid convergence in-distribution regimes, their performance degrades under unmodeled dynamics and distribution shifts. In contrast, bounded ES guarantees robustness to unknown and drifting control directions, though at the cost of slower convergence. By combining these approaches through a safety-aware supervisor and warm-starting ES from DRL actions, the proposed controller maintained high rewards under severe perturbations and outperformed either method alone. These results suggest a principled path toward deploying learning-based controllers in high-dimensional, safety-critical applications such as particle accelerators and robotic arms, where adaptability and robustness are essential.


\bibliographystyle{IEEEtran}
\bibliography{mybibfile}
\nocite{}

\end{document}